\documentclass[]{article}
\usepackage[letterpaper]{geometry}
\usepackage{mtsummit2017}
\usepackage{times}
\usepackage{url}
\usepackage{latexsym}
\usepackage{natbib}
\usepackage{layout}
\usepackage{graphicx}
\usepackage{float}
\usepackage{placeins}
\usepackage{dblfloatfix}
\usepackage{wrapfig}
\usepackage{afterpage}
\usepackage[normalem]{ulem}
\usepackage{color}
\usepackage[utf8]{inputenc}

\def\Sref#1{Section~\ref{#1}} 
 
\def\Fref#1{Figure~\ref{#1}} 
 
\def\Tref#1{Table~\ref{#1}}

\def\equo#1{``#1''}

\begin{document}

\title{Paying Attention to Multi-Word Expressions in Neural Machine Translation}  

\author{\name{\bf Matīss Rikters} \hfill  \addr{matiss@lielakeda.lv}\\ 
		\addr{Faculty of Computing, University of Latvia}\\
\AND
        \name{\bf Ondřej Bojar} \hfill \addr{bojar@ufal.mff.cuni.cz}\\ 
        \addr{Charles University, Faculty of
Mathematics and Physics, \\ Institute of Formal and Applied Linguistics}
}

\maketitle
\pagestyle{empty}

\begin{abstract}
Processing of multi-word expressions (MWEs) is a known problem for any natural
language processing task. Even neural machine translation (NMT) struggles to
overcome it. This paper presents results of experiments on investigating NMT
attention allocation to the MWEs and improving automated translation of
sentences that contain MWEs in English\(\rightarrow\)Latvian and
English\(\rightarrow\)Czech NMT systems. Two improvement strategies were
explored---(1) bilingual pairs of automatically extracted MWE candidates were
added to the parallel corpus used to train the NMT system, and (2) full
sentences containing the automatically extracted MWE candidates were added to
the parallel corpus. Both approaches allowed to increase automated evaluation
results. The best result---0.99 BLEU point increase---has been reached with the
first approach, while with the second approach minimal improvements achieved. We
also provide open-source software and tools used for MWE extraction and
alignment inspection.
\end{abstract}

\section{Introduction}
It is well known that neural machine translation (NMT) has defined the new state
of the art in the last few years \citep{sennrich2016edinburgh,wu2016google}, but
the many specific aspects of NMT outputs are not yet explored. One of which is
translation of multi-word units or multi-word expressions (MWEs). MWEs are
defined by \citet{baldwin2010multiword} as ``lexical items that: (a) can be
decomposed into multiple lexemes; and (b) display lexical, syntactic, semantic,
pragmatic and/or statistical idiomaticity". MWEs have been a challenge for
statistical machine translation (SMT). Even if standard phrase-based models can copy MWEs
verbatim, they suffer in grammaticality. NMT, on the other hand, may struggle in memorizing and reproducing MWEs,
because it represents the whole sentence in a high-dimensional
vector, which can lose the specific meanings of the MWEs even in the more fine-grained attention model \citep{bahdanau:etal:attention:2014}, because MWEs may not appear frequently enough
in the training data.

The goal of this research is to examine how MWEs are treated by NMT systems,
compare that with related work in SMT, and find ways to improve MWE translation
in NMT. We aimed to compare how NMT pays attention to MWEs during translation,
using a test set particularly targeted at handling of MWEs, and if that can be
improved by populating the training data for the NMT systems with parallel
corpora of MWEs.

The objective was to obtain a comparison of how NMT with regular training data
and NMT with synthetic MWE data pays attention to MWEs during the translation
process as well as to improve the final NMT output. To achieve this objective,
it needed to be broken down into smaller sub-objectives: 
\begin{itemize}
\item Train baseline NMT systems,
\item Extract parallel MWE corpora from the training data,
\item Train the NMT systems with synthetic MWE data, and
\item Inspect alignments produced by the NMT.
\end{itemize}

The structure of this paper is as follows: \Sref{related} summarizes related work in
translating MWEs with SMT and NMT. \Sref{data-and-systems} describes the architecture of the
baseline system and outlines the process of extracting parallel MWE corpora from
the training data. \Sref{exps} provides the experiment setup and results. Finally,
conclusions and aims for further directions of work are summarized in \Sref{conc}.

\section{Related Work}
\label{related}

There have been several experiments with incorporating separate processing of
MWEs in rule-based  \citep{deksne2008dictionary} and statistical machine
translation tasks \citep{bouamor2012identifying,skadina2016multi}. However,
there is little literature about similar integrations in NMT workflows so far.

\citet{skadina2016multi} performed a series of experiments on extracting MWE
candidates and integrating them in SMT. The author experimented with several
different methods for both the extraction of MWEs and integration of the
extracted MWEs into the MT system. In terms of automatic MT evaluation, this
allowed to achieve an increase of ~0.5 BLEU points \citep{papineni2002bleu} for
an English\(\rightarrow\)Latvian SMT system.

\citet{tang2016neural} introduce an NMT approach that uses a stored phrase
memory in symbolic form. The main difference from traditional NMT is tagging
candidate phrases in the representation of the source sentence and forcing the
decoder to generate multiple words all at once for the target phrase. Although
they do mention MWEs, no identification or extraction of MWEs is performed and
the phrases they mainly focus on are dates, names, numbers, locations, and
organizations, that are collected from multiple dictionaries. For
Chinese\(\rightarrow\)English they report a 3.45 BLEU point increase over
baseline NMT.

\citet{cohn2016incorporating} describe an extension of the traditional
attentional NMT model with the inclusion of structural biases from word-based
alignment models, such as positional bias, Markov conditioning, fertility and
agreement over translation directions. They perform experiments translating
between English, Romanian, Estonian, Russian and Chinese and analyze the
attention matrices of the output translations produced by running experiments
using the different biases. Specific experiments targeting MWEs are not
performed, but they do point out that using fertility, especially global
fertility, can be useful for dealing with multi-word expressions. They report a
statistically significant improvement of BLEU scores in almost all involved
language pairs.

\citet{chen:etal:quided:ali:2016} use a similar approach as we do. Their ``bootstrapping''  automatically extracts smaller parts of training segment pairs and adds them to the training data for NMT. The main difference is that they rely on automatic word alignment and punctuation in the sentence to identify matching sub-segments. 

\section{Data Preparation and Systems Used}
\label{data-and-systems}

To measure changes introduced by adding synthetic MWE data to the training
corpora, first, a baseline NMT system was trained for each language pair. The
experiments were conducted on English\(\rightarrow\)Czech and
English\(\rightarrow\)Latvian translation directions.

\subsection{Baseline NMT System}

To be able to compare the results with other MT systems, training and
development corpora were used from the WMT  shared tasks: data from the News
Translation Task\footnote{\url{http://www.statmt.org/wmt17/translation-task.html}} for English\(\rightarrow\)Latvian and data from the Neural MT
Training Task\footnote{\url{http://www.statmt.org/wmt17/nmt-training-task/}} \citep{wmt17ntt:2017} for English\(\rightarrow\)Czech. The English\(\rightarrow\)Czech
data consists of about 49 million parallel sentence pairs and the
English\(\rightarrow\)Latvian of about 4.5 million. The development corpora
consist of 2003 sentences for English\(\rightarrow\)Latvian and 6000 for
English\(\rightarrow\)Czech. 

Neural Monkey \citep{NeuralMonkey:2017}, an open-source tool for sequence learning, was used to train the baseline NMT systems. Using the configuration provided
by the WMT Neural MT Training Task organizers, the baseline reached 11.29 BLEU points for
English\(\rightarrow\)Latvian after having seen 23 million sentences in about 5
days and 13.71 BLEU points for English\(\rightarrow\)Czech after having seen 18
million sentences in about 7 days.

\subsection{Extraction of Parallel MWEs}

To extract MWEs, the corpora were first tagged with morphological taggers:
UDPipe  \citep{ramisch2012generic} for English and Czech, LV Tagger
\citep{paikens2013morphological} for Latvian. After that, the tagged corpora were
processed with the Multi-word Expressions toolkit  \citep{ramisch2012generic},
and finally aligned with the MPAligner \citep{pinnis2013context}, intermittently
pre-processing and post-processing with a set of custom tools. To extract MWEs
from the corpora with the MWE Toolkit, patterns were required for each of the
involved languages. Patterns from \citet{skadina2016multi} were used for Latvian
(210 patterns) and English (57 patterns) languages and patterns from
\citet{majchrakovasemi} and \citet{pecina2008reference} for Czech (23 patterns).

This workflow allowed to extract a parallel corpus of about 400 000 multi-word
expressions for English\(\rightarrow\)Czech and about 60 000 for
English\(\rightarrow\)Latvian. For an extension of this experiment, all
sentences containing these MWEs were also extracted from the training corpus, serving as a separate parallel
corpus.

\section{Experiments}
\label{exps}

\begin{figure*}[t]
  \includegraphics[width=\linewidth]{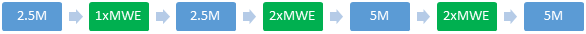}
  \caption{Portions of the final training data set for English\(\rightarrow\)Czech}
  \label{fig:en-cs-portions}
\end{figure*}
\begin{figure*}[t]
  \includegraphics[width=\linewidth]{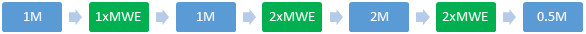}
  \caption{Portions of the final training data set for English\(\rightarrow\)Latvian}
  \label{fig:en-lv-portions}
\end{figure*}

We experiment with two forms of the presentation of MWEs to the NMT system: (1)
we add only the parallel MWEs themselves, each pair forming a new ``sentence
pair'' in the parallel corpus, and (2) we use full sentences containing the
MWEs. We denote the approaches \equo{MWE phrases}  and \equo{MWE sents.} in the
following.

\subsection{Training Corpus Layout}

In both cases, we use the same corpus training corpus layout: we mix the
baseline parallel corpus with synthetic data so that MWEs 
%
get more exposure to the
neural network in training and hopefully allow NMT to learn to translate them
better.

\Fref{fig:en-cs-portions} and \Fref{fig:en-lv-portions} illustrate
how the training data was divided into portions. The block 1xMWE corresponds to
the
full set of extracted MWEs (400K for En\(\rightarrow\)Cs, 60K for
En\(\rightarrow\)Lv) and 2xMWE corresponds to two copies  of the set (800K for
En\(\rightarrow\)Cs, 120K for En\(\rightarrow\)Lv). For En\(\rightarrow\)Lv the
full corpus was used. For En\(\rightarrow\)Cs we used only the first 15M
sentences to be able to train multiple epochs on the available
hardware. The MWEs get repeated five times in both language pairs.
By doing this, the En\(\rightarrow\)Cs data set was reduced from 49M to 17M and
the En\(\rightarrow\)Lv data set increased to 4.8M parallel sentences for one epoch of training.

While the experiments were running, early stopping of the training was executed
and snapshots of the models for evaluation were taken in stages where the models
already were starting to converge. For En\(\rightarrow\)Lv this was after the
networks had been trained on 25M sentences (i.e. 5.2~epochs of the mixed corpus), for En\(\rightarrow\)Cs 27M
sentences (i.e. 1.6~epochs).


Neural Monkey does not shuffle the
training corpus between epochs. This is not a problem if the corpus is properly
shuffled and the number of epochs is not very large compared to the size of the
epochs. We shuffled only the baseline corpus and the interleaved it with
(shuffled) sections for MWEs. This worked well when MWEs were provided in full
sentences, but not with MWEs presented as expressions. In the latter case, the
NMT started to produce only very short output, losing very much of its
performance. We, therefore, shuffle the whole composed corpus for the
\equo{MWE phrases} runs, effectively discarding the interleaved composition of
the training data.

\subsection{Results}

\Tref{tab:results} shows the results for both approaches and both language
pairs.
Due to hardware constraints, we were not able to try out both approaches on
both language pairs.

We evaluate all setups with BLEU \citep{papineni2002bleu} on the full
development set (distinct from the training set), as shown in the column
\equo{Dev}, and on a subset of 611 (En\(\rightarrow\)Lv) and 112 (En\(\rightarrow\)Cs) sentences containing the
identified MWEs (column \equo{MWE}).



\begin{table}[t]
  \begin{center}
  \begin{tabular}{|l|c|c|c|c|}
  \hline 
  Languages & \multicolumn{2}{|c|}{En\(\rightarrow\)Cs} & \multicolumn{2}{|c|}{En\(\rightarrow\)Lv} \\ 
  \hline 
  Dataset & Dev& MWE  & Dev & MWE \\ 
  \hline
  Baseline & 13.71 & 10.25 & 11.29 & 9.32 \\
  +MWE phrases & - & - & \bf 11.94 & \bf 10.31 \\
  +MWE sents. & \bf 13.99 & \bf 10.44 & - & -  \\
  \hline
  \end{tabular}
  \end{center}
  \caption{ Experiment results. }
  \label{tab:results}
\end{table}

\begin{figure}[t]
\centering
\begin{minipage}{.48\textwidth}
  \centering
  \includegraphics[width=\linewidth]{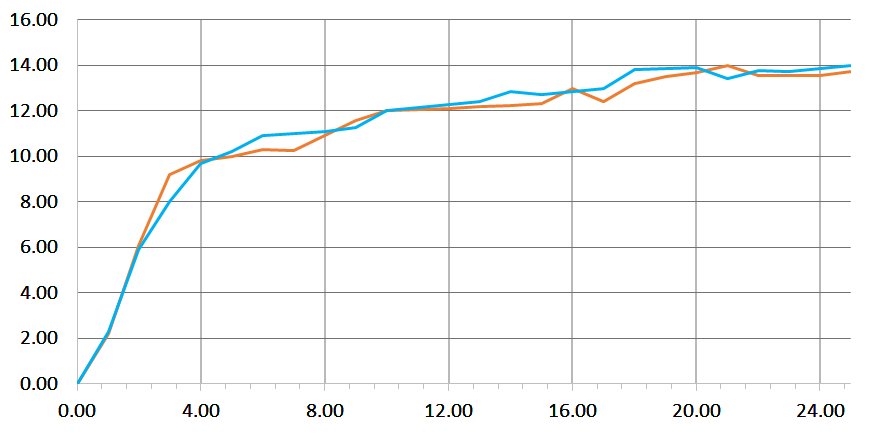}
  \caption{Automatic evaluation progression of En\(\rightarrow\)Cs experiments on validation data. Orange -- baseline; blue –- baseline with added MWEs.}
  \label{fig:bleu-en-cs}
\end{minipage}%
\hfill
\begin{minipage}{.48\textwidth}
  \centering
  \includegraphics[width=\linewidth]{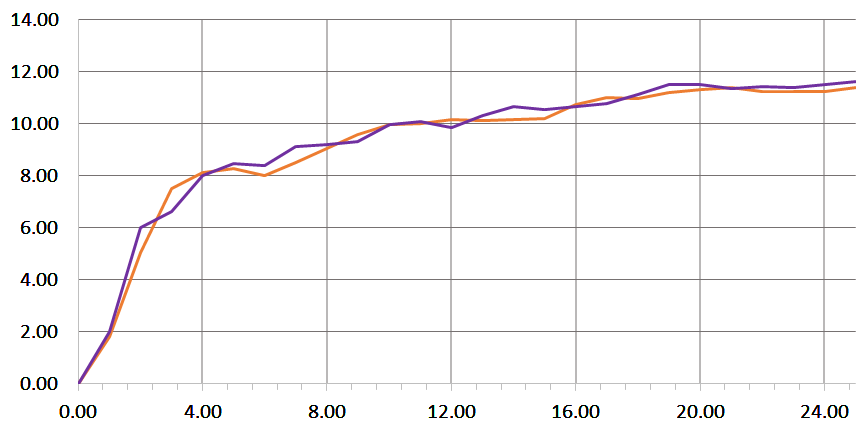}
  \caption{Automatic evaluation progression of En\(\rightarrow\)Lv experiments on validation data. Orange -- baseline; purple –- baseline with added MWE sentences.}
  \label{fig:bleu-en-lv}
\end{minipage}
\end{figure}

Figures \ref{fig:bleu-en-cs} and \ref{fig:bleu-en-lv} illustrate the learning
curves in terms of millions of sentences, as evaluated on the full development set.

We see that the difference on the whole development set is not very big for
either of the languages, and that it fluctuates as the training progresses.

The
improvement is more apparent when evaluated on the dedicated devset of sentences
containing multi-word expressions. The improvement for Latvian is even 0.99
BLEU, but arguably, the baseline performance of our system is not very high.
Also,
more runs should be carried out for a full confidence, but this was
unfortunately out of our limits on computing resources.

\begin{figure*}[t]
  \includegraphics[width=\linewidth]{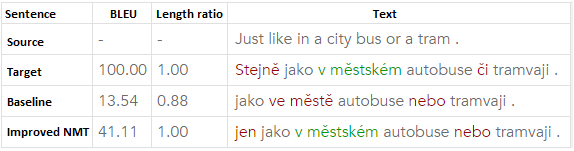}
  \caption{Differences in translation between baseline and improved NMT system. Improving n-grams are highlighted in green and worsening n-grams –- in red.}
  \label{fig:city-bus}
\end{figure*}

\begin{figure*}[t]
  \includegraphics[width=\linewidth]{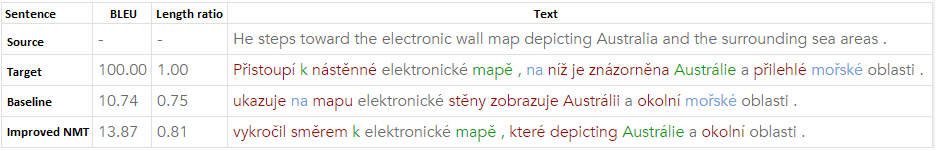}
  \caption{Differences in translation of a Czech sentence using baseline and improved NMT systems. Improving n-grams are highlighted in green and worsening n-grams –- in red.}
  \label{fig:differences}
\end{figure*}

\begin{figure*}[t]
  \includegraphics[width=\linewidth]{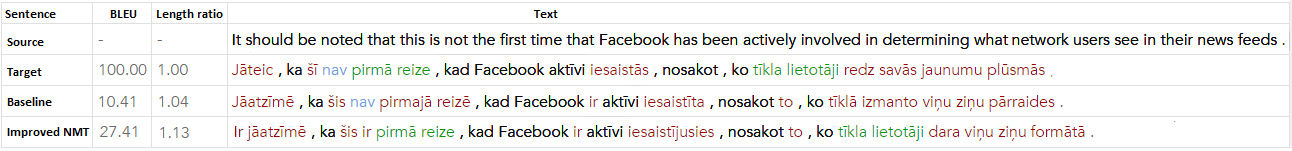}
  \begin{tabular}{lp{10.4cm}}
   \bf Source: & It should be noted that this is not the first time that Facebook has been actively involved in determining what network users see in their news feeds. \\
   \bf Baseline: & Jāatzīmē, ka šis nav pirmajā reizē, kad Facebook ir aktīvi iesaistīta, nosakot to, ko tīklā izmanto viņu ziņu pārraides.\\
   \bf Improved NMT: & Ir jāatzīmē, ka šis ir pirmā reize, kad Facebook ir aktīvi iesaistījusies, nosakot to, ko tīkla lietotāji dara viņu ziņu formātā.\\
   \bf Reference: & Jāteic, ka šī nav pirmā reize, kad Facebook aktīvi iesaistās, nosakot, ko tīkla lietotāji redz savās jaunumu plūsmās. \\
  \end{tabular}
  \caption{Differences in translation between baseline and improved NMT system. Improving n-grams are highlighted in green and worsening n-grams –- in red.}
  \label{fig:network-users}
\end{figure*}



\subsection{Manual Inspection}

To find out whether changes in the results are due to the synthetic MWE corpora
added, a subset of output sentences from the ones containing MWEs were selected
for closer examination. For this task, we used the iBLEU \citep{madnani2011ibleu}
tool.

In \Fref{fig:city-bus}, an improvement in the modified NMT translation is
visible due to the treatment of the compound nominal ``city bus" as a single
expression. It seems that the baseline system translates ``city" into
``m\v{e}st\v{e}" and ``bus" into ``autobuse" individually, resulting in the
wrong form of ``city" in Czech (a noun used instead of an adjective). On the other hand, the improved NMT translates ``city"
into ``m\v{e}stsk\'{e}m" just like the target human translation. Attention
alignments will be examined in the following section.

\Fref{fig:differences} shows an example where the improved NMT scores
higher in BLEU points and translates the MWE closer to the human, but loses a
part of it in the process. While translating the noun phrase ``electronic wall
map" the improved system generates a closer match to the human translation
``elektronick\'{e} map\v{e}", it does not translate the word ``wall" that was
translated into ``st\v{e}ny" by the baseline system. Upon closer inspection, we
discovered that this error was caused by the MWE extractor and aligner because
the identified English phrase ``electronic wall map" was aligned to an
identified Czech phrase ``elektronick\'{e} map\v{e}" and the whole phrase
``n\'{a}st\v{e}nn\'{e} elektronick\'{e} map\v{e}" was not identified by the MWE
extractor at all.

Figure \ref{fig:network-users} illustrates translations of an example sentence
by the En\(\rightarrow\)Lv NMT systems. The MWE, in this case, is ``network users"
that is translated as ``t\={\i}kla lietot\={a}ji" by the modified system and
completely mistranslated by the baseline.

\subsection{Alignment Inspection}

\begin{wrapfigure}{r}{0.5\textwidth}
  \begin{center}
    \includegraphics[width=0.4\textwidth]{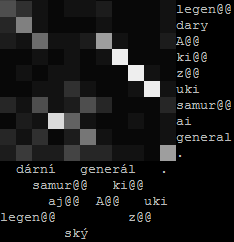}
  \end{center}
  \caption{Example of a soft alignment matrix.}
  \label{fig:matrix}
\end{wrapfigure}

\begin{figure}[t]
\centering
\begin{minipage}{.48\textwidth}
  \centering
  \includegraphics[width=\linewidth]{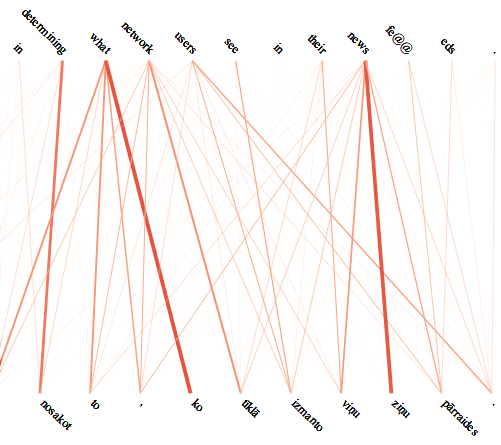}
  \caption{Fragment of soft alignments of the example sentence from the baseline NMT system.}
  \label{fig:bad-Facebook}
\end{minipage}%
\hfill
\begin{minipage}{.48\textwidth}
  \centering
  \includegraphics[width=\linewidth]{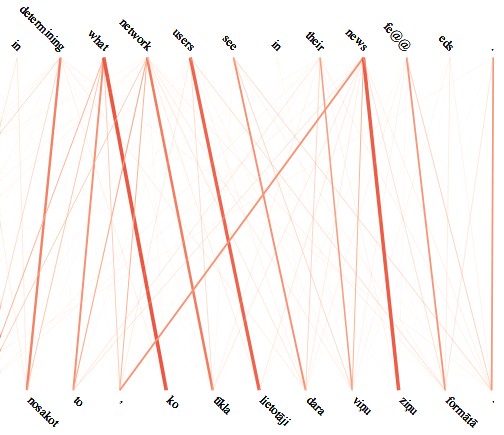}
  \caption{Fragment of soft alignments of the example sentence from the improved NMT system.}
  \label{fig:good-Facebook}
\end{minipage}
\end{figure}

For inspecting the NMT attention alignments, we developed a tool\footnote{NMT Attention Alignment Visualizations - \url{https://github.com/M4t1ss/SoftAlignments}}
\citep{attention:visualization:tool:2017} that takes data
produced by Neural Monkey---a 3D array (tensor) filled with the alignment
probabilities together with source and target subword units
\citep{sennrich2015neural} or byte pair encodings (BPEs)---as input and produces a
soft alignment matrix (\Fref{fig:matrix}) of the subword units that
highlights all units, that get attention when translating a specific subword
unit. The tool includes a web version that was adapted from Nematus
\citep{sennrich2017nematus} utilities and slightly modified. It allows to output
the soft alignments in a different perspective, as connections between BPEs as
visible in \Fref{fig:bad-Facebook} and  \Fref{fig:good-Facebook}.

\afterpage{\clearpage}

\begin{figure*}[t]
  \includegraphics[width=\linewidth]{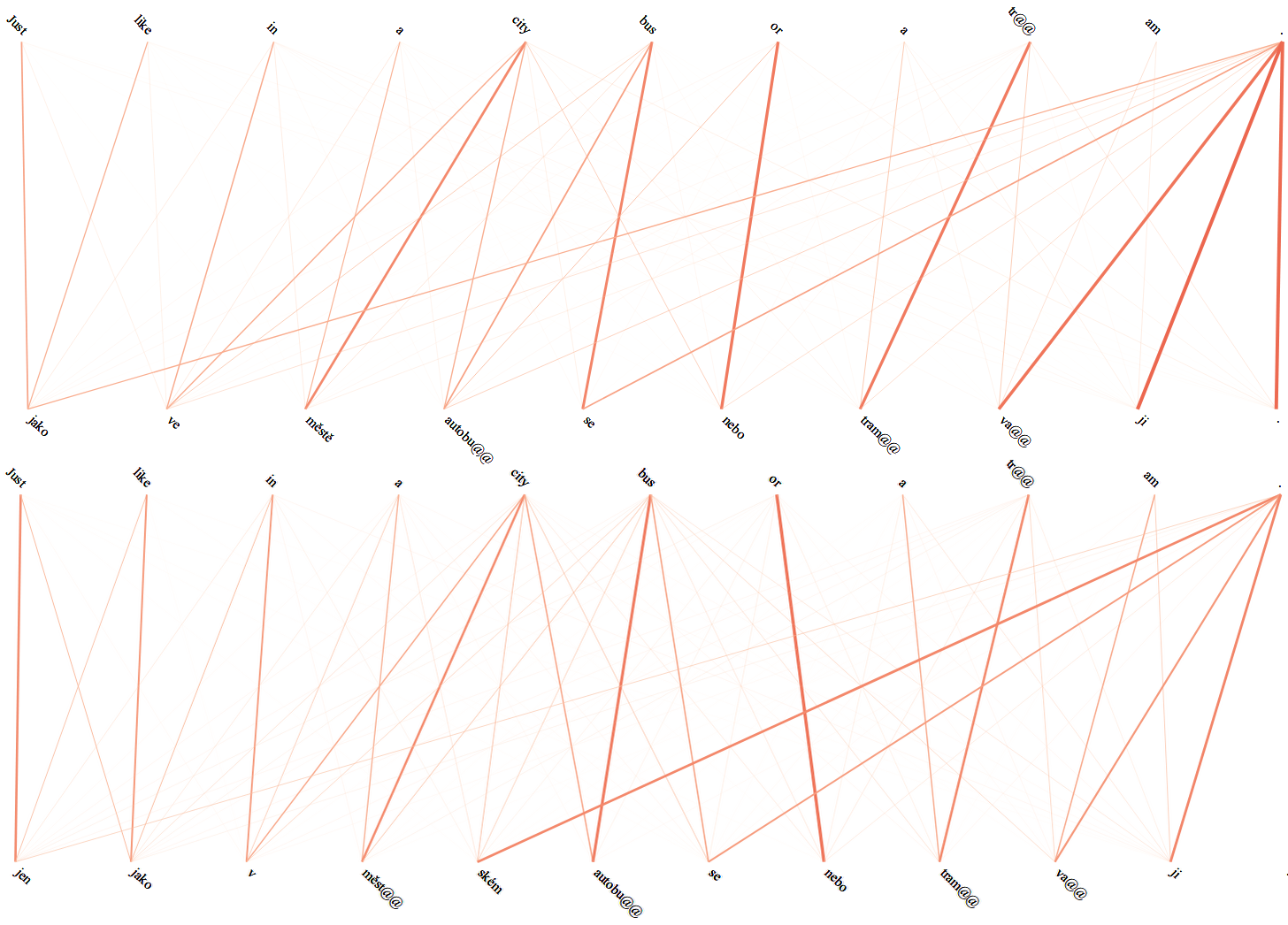}
  \begin{tabular}{lp{10.4cm}}
   \bf Source: & Just like in a city bus or a tram. \\
   \bf Baseline: & Jako ve městé autobuse nebo tramvaji.\\
   \bf Improved NMT: & Jen jako v městském autobuse nebo tramvaji.\\
   \bf Reference: & Stejně jako v městském autobuse či tramvaji. \\
  \end{tabular}
  \caption{Soft alignment example visualizations from translating an English sentence into Czech from the baseline (top, hypothesis 1) and improved (bottom, hypothesis 2) NMT systems.}
  \label{fig:city-bus-ali}
\end{figure*}

In these examples, the attention state of the previously mentioned MWE from
En\(\rightarrow\)Lv translations (``network users") is visible. The
alignment inspection tool allows to see that the baseline NMT in  \Fref{fig:bad-Facebook} has multiple faded alignment lines for both words
``network" and ``users", which outlines that the neural network is unsure and
looking all around for traces to the correct translation. However, in
\Fref{fig:good-Facebook}, it is visible that both these words have strong
alignment lines to the words ``t\={\i}kla lietot\={a}ji", that were also
identified by the MWE Toolkit as an MWE candidate.

Figure \ref{fig:city-bus-ali} shows one of the previously mentioned
En\(\rightarrow\)Cs translation examples. Here it is clear that in the baseline
alignment no attention goes to the word ``m\v{e}st\v{e}" or the subword units
``autobu@@" and ``se" when translating ``city". In the modified version, on the
other hand,
some attention from ``city" goes into all closely related subword units:
``m\v{e}st@@", ``sk\'{e}m", ``autobu@@", and ``se". It is also visible that in
this example, the translation of ``bus" gets attention from not only ``autobu@@"
and ``se" but also the ending subword unit of ``city", i.e. the token ``sk\'{e}m".

\FloatBarrier

\section{Conclusion}
\label{conc}

In this paper, we described the first experiments with handling multi-word
expressions in neural machine translation systems. Details on identifying and
extracting MWEs from parallel corpora, as well as aligning them and building
corpora of parallel MWEs were provided. We explored two methods of integrating
MWEs in training data for NMT and examined the output translations of the
trained NMT systems with custom built tools for alignment inspection.

In addition to the methods described in this paper, we also released open-source
scripts\footnote{Multiword-Expression-Tools - \url{https://github.com/M4t1ss/MWE-Tools}} for a complete workflow of identifying, extracting and integrating MWEs
into the NMT training and translation workflow.

While the experiments did not show outstanding improvements on the general
development data set, an increase of 0.99 BLEU was observed when using an MWE
specific test data set. Manual inspection of the output translations confirmed
that translations of specific MWEs were improving after populating the training
data with synthetic MWE data.

As the next steps, we plan (1) to analyze the obtained results of our
experiments in more detail through the help of a larger scale manual human
evaluation of the NMT output and (2) to continue experiments to find best ways
how to treat different categories of MWEs, i.e. idioms.

\section*{Acknowledgement}
This study was supported in parts by the grants H2020-ICT-2014-1-645442 (QT21), the ICT COST Action IC1207 \emph{ParseME: Parsing and multi-word expressions. Towards linguistic precision
and computational efficiency in natural language processing}, 
 and Charles University Research Programme ``Progres'' Q18 -- Social Sciences: From Multidisciplinarity to Interdisciplinarity.

\FloatBarrier

\bibliographystyle{apalike}
\bibliography{mybib}

\end{document}